\documentclass{article}
\usepackage{amsmath,graphicx,mlspconf}
\usepackage{booktabs}
\usepackage{xcolor}
\usepackage[hyphens]{url}
\usepackage{hyperref}
\hypersetup{breaklinks=true, hidelinks}
\usepackage{soul}
\usepackage{balance}

\copyrightnotice{%
\parbox{7in}{\footnotesize
{\copyright} 2026 IEEE. Personal use of this material is permitted. Permission from IEEE must be obtained for all other uses, in any current or future media, including reprinting/republishing this material for advertising or promotional purposes, creating new collective works, for resale or redistribution to servers or lists, or reuse of any copyrighted component of this work in other works.}%
}

\toappear{2026 IEEE International Workshop on Machine Learning for Signal Processing, Sep.\ 28-- Oct.\ 1, 2026, Atlanta, USA}

\title{BEAT-SYNCHRONOUS TOKENIZATION FOR ECG TRANSFORMERS}

\name{%
Ahmed Sameh$^{1}$,
Nolan Wilson$^{1}$,
Max Enderlein$^{1}$,
Yogatheesan Varatharajah$^{1}$%
\thanks{{\small $^{1}$ Computer Science \& Engineering,
University of Minnesota Twin Cities, Minneapolis, MN 55455, USA.
{\tt \{sameh002, wils2744, ender121, yvaratha\}@umn.edu}}}%
}
\address{}
\begin{document}
\maketitle

\begin{abstract}
Transformer-based electrocardiogram (ECG) models commonly tokenize waveforms into fixed temporal patches. Though convenient, fixed patching can split heartbeat structures across token boundaries. We study beat-synchronous tokenization as a physiologically grounded alternative, comparing fixed patches with three beat-aligned strategies: resampled beats, adaptive pooled beats, and resampled beats augmented with R--R interval information. Experiments span two settings: 10-second 12-lead diagnostic classification on PTB-XL after MIMIC-IV-ECG masked pretraining, and 60-second single-lead rhythm classification on Icentia11k after patient-level contrastive pretraining. On PTB-XL, resampled beat tokens achieve the highest mean macro Area Under the ROC Curve (AUROC; 0.8945) and nearly match the best fixed-patch macro Area Under the Precision-Recall Curve (AUPRC; 0.7414), reducing average sequence length from 100 to 11.2 tokens. On Icentia11k, beat-synchronous tokenizers obtain comparable AUPRC to fixed patching with better stability across runs. These results suggest morphology-preserving beat tokenization is a compact, competitive alternative to fixed temporal patching.
\end{abstract}


\section{Introduction}
\label{sec:intro}

Electrocardiography (ECG) provides a compact, non-invasive view of cardiac electrical activity and remains central to cardiovascular screening, diagnosis, and monitoring. Deep learning has substantially improved automated ECG analysis across clinical prediction tasks \cite{attia2019nm_contractile,attia2019lancet_af,hannun2019cardiologist}. However, fully supervised ECG models often depend on large expert-labeled datasets, which are costly to curate and difficult to scale across institutions, devices, and patient populations. Self-supervised learning (SSL) addresses this bottleneck by pretraining ECG encoders on unlabeled recordings and transferring the learned representations to downstream tasks \cite{diamant2022pclr,mehari2021selfsupervised_12lead}. This direction has recently expanded toward ECG foundation models trained at large scale for broad transfer across clinical settings \cite{mckeen2024ecgfm,li2024ecgfounder}.

Despite these advances, many Transformer-based ECG models still represent the waveform as a sequence of fixed temporal patches. Fixed patching is simple and compatible with common SSL objectives such as masked reconstruction \cite{na2024guiding} or contrastive learning \cite{kiyasseh2021clocs,diamant2022pclr}, but it treats ECGs as generic time series rather than structured cardiac recordings. A fixed patch may split a cardiac cycle across token boundaries or mix adjacent beats, even though ECG interpretation often depends on beat morphology, inter-beat timing, and rhythm-level organization. Since heart rate varies across patients and recording conditions, the same patch length can correspond to different physiological content across examples.

Recent ECG models have begun to replace fixed temporal patches with beat-aware or physiologically structured representations \cite{jin2025reading,wang2026rhythmbert,10752959,ma2026beat,pmlr-v267-wang25du}. These studies support the value of aligning ECG representations with cardiac structure, but tokenization is often introduced together with other modeling changes, such as specialized architectures, discrete vocabularies, or ECG-text training. However, the effect of tokenization alone on performance and efficiency remains unclear.

In this work, we test the hypothesis that beat-synchronous tokens are more efficient alternatives to fixed temporal patches under matched Transformer pretraining and downstream evaluation protocols. We compare fixed temporal patches with three beat-synchronous tokenizers: resampled beat tokens, adaptive pooled beat tokens, and resampled beat tokens augmented with R--R interval information. This setup allows us to evaluate whether cardiac-cycle alignment improves the performance--efficiency tradeoff while keeping the encoder family and evaluation protocol controlled.

We evaluate our hypothesis in two complementary settings. First, we pretrain 12-lead ECG Transformers on MIMIC-IV-ECG \cite{PhysioNet-mimic-iv-ecg-1.0} using the masked reconstruction SSL objective and evaluate transfer to PTB-XL \cite{PhysioNet-ptb-xl-1.0.3} diagnostic classification on five superclasses. Second, we pretrain single-lead models on the Icentia11k dataset using a contrastive learning objective and evaluate 60-second dominant rhythm classification \cite{tan2019icentia11k}. These settings test beat-synchronous tokenization on both standard 10-second clinical ECGs and longer-term ambulatory recordings.

Our results show that beat-synchronous tokenization can retain strong downstream performance while using substantially shorter token sequences. On PTB-XL, the resampled beat tokenizer achieves the highest mean macro AUROC, while the R--R augmented tokenizer remains close to the strongest fine-patch baseline; both use only 11.2 tokens on average compared with 100 tokens for the finest fixed-patch tokenizer. The adaptive-pooling variant performs substantially worse, indicating that beat alignment alone is insufficient without a morphology-preserving beat encoder. On Icentia11k, where atrial fibrillation/atrial flutter (AFib/AFL) windows are rare, beat-synchronous tokenizers obtain comparable AUPRC to fixed patching and show lower run-to-run AUPRC variability, while using approximately 68 beat tokens on average compared with 93 fixed tokens. Overall, these findings suggest that beat-synchronous tokenization can be an effective and token-efficient alternative to fixed temporal patching, but the design of the beat encoder is critical.

\section{Related Work}
\label{sec:related}

\noindent\textbf{Self-supervised ECG representation learning:}
SSL has become a common strategy for learning ECG representations from large unlabeled datasets. Contrastive approaches define positive pairs across time, leads, or patients to encourage clinically useful invariances \cite{kiyasseh2021clocs,diamant2022pclr}, while reconstruction-based methods train encoders to recover masked ECG segments or patches \cite{na2024guiding}. Benchmarks on 12-lead ECGs show that SSL can approach the performance of fully supervised learning with reduced label dependence \cite{mehari2021selfsupervised_12lead}, and recent ECG foundation models scale pretraining to larger datasets and broader transfer settings \cite{mckeen2024ecgfm,li2024ecgfounder}. Most of these pipelines, however, represent the waveform using fixed temporal windows or patches, which are convenient for Transformer encoders but not explicitly aligned with cardiac cycles.

\noindent\textbf{ECG tokenization and ECG-language modeling:}
Recent work has explored more structured ECG tokenization. HeartLang treats heartbeats as words and rhythms as sentences, using QRS-aligned ECG sentences and vocabulary-based pretraining \cite{jin2025reading}. RhythmBERT further develops this idea by tokenizing P, QRS, and T waveform components into symbolic representations \cite{wang2026rhythmbert}. These approaches support the idea that ECGs contain natural physiological units that may be better suited to sequence modeling than arbitrary time patches. At the same time, their tokenization choices are embedded within larger systems involving discrete vocabularies, clustering, wave segmentation, or specialized pretraining objectives. Our work is complementary: rather than proposing a full ECG-language framework, we directly compare fixed temporal patches against beat-synchronous continuous tokenizers under matched Transformer settings.

\noindent\textbf{Beat-aware and rhythm-focused modeling:}
Beat-level representations have also been studied in supervised and rhythm-monitoring contexts. Patient-adaptive beat-wise Transformers use beat tokens with patient-specific symbolic morphology for atrial fibrillation detection in long-term monitoring \cite{10752959}. BEAT-Net uses QRS-aligned tokens in a supervised model designed to model morphology, lead-specific spatial information, and temporal rhythm structure \cite{ma2026beat}. MELP incorporates beat-level information into ECG-text pretraining through token-, beat-, and rhythm-level alignment \cite{pmlr-v267-wang25du}. While these studies indicate that beat-level structure is useful, they do not establish whether beat-synchronous token boundaries alone improve the performance--efficiency tradeoff.

\noindent\textbf{Position of this work:}
Prior work motivates physiologically structured ECG representations, but the effect of tokenization is often entangled with other modeling choices. We therefore focus on a controlled comparison: fixed temporal patches versus beat-synchronous tokens, using comparable Transformer encoders, SSL objectives, and downstream evaluation protocols. By testing resampled beats, adaptive pooled beats, and R--R augmented beats, we not only evaluate whether beat alignment helps, but also which beat-token designs preserve morphology and rhythm information effectively.

\section{Methodology}
\label{sec:method}

\begin{figure}[!h]
    \centering
    \includegraphics[width=\linewidth]{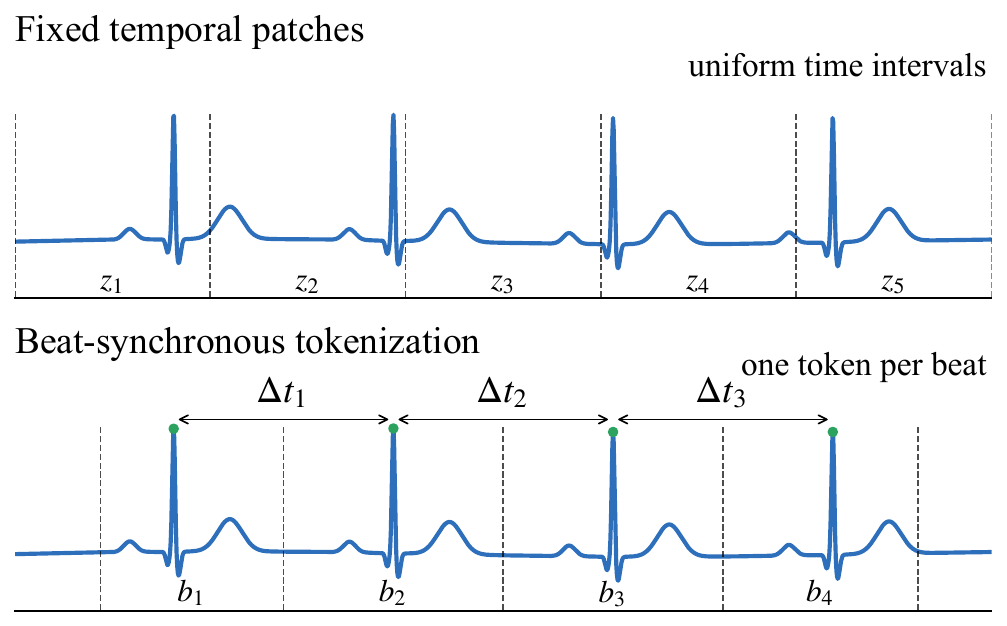}
    \caption{
    Fixed temporal patching divides the ECG into uniform time intervals, whereas beat-synchronous tokenization defines tokens from cardiac cycles identified by R-peaks. 
    $\Delta t$ denotes the R--R interval between consecutive R-peaks, which is added as timing information in Tok3.
    }
    \label{fig:overview}
\end{figure}

Fig.~\ref{fig:overview} summarizes the main difference between fixed temporal patching and beat-synchronous tokenization. We study ECG tokenization as an isolated design choice by keeping the Transformer-style encoder and downstream evaluation protocol comparable across tokenizers. Given an ECG segment $X$, each tokenizer maps the waveform to a token sequence $\{z_i\}_{i=1}^{N}$, which is processed by a Transformer encoder and pooled into a segment-level representation. We compare fixed temporal patches with three beat-synchronous tokenizers.

\noindent\textbf{Fixed temporal patch tokenizer:}
The fixed tokenizer partitions the ECG into non-overlapping temporal patches of length $p$ and embeds each patch using a one-dimensional convolution with kernel size and stride $p$. For 10-second 12-lead ECGs sampled at 500 Hz, we evaluate $p \in \{50,100,250,500\}$, corresponding to 100, 50, 20, and 10 tokens per recording. For the 60-second Icentia11k setting sampled at 250 Hz, we use $p=160$, yielding 93 fixed tokens per window.

\noindent\textbf{Beat-synchronous tokenizers:}
Beat-synchronous tokenization uses consecutive R-peaks to define cardiac-cycle tokens. Let $r_i$ and $r_{i+1}$ be adjacent R-peak locations. The $i$-th beat is extracted from the interval $[r_i,r_{i+1})$, so the token sequence length is determined by the number of detected beats rather than a fixed time grid. We evaluate three beat-token designs.

\noindent\ul{\emph{Tok1 - resampled beat tokens:}} Each variable-length beat is resampled to a fixed length and embedded by a one-dimensional convolution. In the 12-lead MIMIC/PTB-XL setting, each beat is represented as a $12 \times 300$ segment. In the single-lead Icentia setting, each beat is represented as a $1 \times 160$ segment.

\noindent\ul{\emph{Tok2 - adaptive pooled beat tokens:}} Tok2 preserves beat boundaries but avoids explicit temporal resampling. Variable-length beat segments are passed through a convolutional front end and compressed to one token using adaptive average pooling. This tokenizer tests whether beat alignment alone is sufficient when beat morphology is aggressively compressed.

\noindent\ul{\emph{Tok3 - resampled beat tokens with R--R timing:}} Tok3 uses the same resampled beat representation as Tok1, but additionally embeds the R--R interval associated with each beat using a small multilayer perceptron. The R--R embedding is added to the beat embedding before the Transformer encoder, providing explicit rhythm timing information.

\noindent\textbf{Encoder architecture:}
All tokenizers are paired with Transformer encoders using hidden dimension $d=256$, 8 attention heads, GELU activations, pre-layer normalization, and dropout 0.1. For the MIMIC/PTB-XL experiments, we use 4 Transformer layers. For the Icentia experiments, we use 6 Transformer layers. Fixed-patch encoders use positional encodings over fixed token sequences. Beat-synchronous encoders use padding masks because the number of beats varies across recordings; valid token representations are mean-pooled after the Transformer layers.

\noindent\textbf{Self-supervised objectives:}
We use the SSL objective that best matches each experimental setting: masked reconstruction for 10-second 12-lead diagnostic transfer, where preserving waveform morphology is central, and patient-level contrastive learning for Icentia11k, where long ambulatory recordings allow positive pairs to be sampled from different windows of the same patient. SSL models were trained until convergence, not for a fixed budget, so differences between tokenizers were not driven by unequal pretraining progress.

\noindent\textbf{Masked reconstruction pretraining for 12-lead ECG:}
For the MIMIC-IV-ECG experiments, we pretrain each tokenizer using masked reconstruction. A subset of tokens is replaced by a mask token, the Transformer processes the masked sequence, and a prediction head reconstructs the waveform patch or beat segment. Fixed-patch models reconstruct temporal patches of dimension $12p$, while beat-synchronous models reconstruct beat-level targets. We use a 0.5 mask ratio throughout. Pretraining uses AdamW with learning rate $2\times10^{-5}$, weight decay 0.05, and batch size 256.

\noindent\textbf{Patient-level contrastive pretraining for Icentia11k:}
We pretrain a 60-second single-lead fixed-patch encoder using patient-level contrastive learning. Each positive pair consists of two different 60-second windows sampled from the same patient, while windows from other patients in the minibatch are negatives. The encoder is trained with an InfoNCE-style contrastive objective \cite{oord2018representation}. In the downstream Icentia comparison, the fixed-patch model uses this pretrained encoder directly. The beat-synchronous and beat-synchronous+HR models initialize shared convolutional and Transformer weights from the same fixed-patch checkpoint by applying the learned 160-sample convolution to resampled beat windows; Tok3 additionally learns the R--R interval encoder during downstream training. This design makes the experiment a direct comparison of downstream tokenization choices under the same pretrained initialization. We evaluate Tok1 and Tok3 in this setting because Tok2 performed substantially worse in the PTB-XL experiment, indicating that adaptive pooling was not a competitive beat-token design.

\section{Experimental Setup}
\label{sec:experiments}

\noindent\textbf{MIMIC-IV-ECG pretraining data:}
We use MIMIC-IV-ECG as the unlabeled 12-lead pretraining corpus \cite{PhysioNet-mimic-iv-ecg-1.0}. Records are represented as 10-second, 12-lead ECGs sampled at 500 Hz, giving tensors of shape $12 \times 5000$. Invalid values are sanitized by removing highly corrupted records, interpolating minor missing segments, clipping amplitudes to $[-5,5]$, and applying per-lead z-score normalization. For beat-synchronous pretraining, R-peaks are detected from lead II, and consecutive R-peaks define beat intervals. We use a maximum of 35 beats for padding and positional encoding.

\noindent\textbf{PTB-XL downstream task:}
We evaluate 12-lead transfer on PTB-XL five-superclass diagnostic classification \cite{PhysioNet-ptb-xl-1.0.3}. The Standard Communication Protocol (SCP) codes are mapped to the standard diagnostic superclasses: NORM (Normal ECG), MI (Myocardial Infarction), STTC (ST/T-Change), CD (Conduction Disturbance), and HYP (Hypertrophy). We follow the official PTB-XL stratified split: folds 1--8 for training, fold 9 for validation, and fold 10 for testing. Records without any superclass label or missing waveform files are removed. Each model is fine-tuned for multi-label classification with a five-output prediction head. We select checkpoints by validation macro AUPRC and report test macro AUROC and macro AUPRC over five runs.

\noindent\textbf{PTB-XL token counts:}
For fixed patches, the token count is determined by the patch size: $p=50$ gives 100 tokens, $p=100$ gives 50 tokens, $p=250$ gives 20 tokens, and $p=500$ gives 10 tokens. For beat-synchronous tokenizers, the sequence length is determined by detected cardiac cycles. On the PTB-XL training set, beat tokenization yields 11.2 beats per record on average, with median 11, interquartile range 10--12, 90th percentile 14, and 95th percentile 16. The maximum padding length of 35 covers all records.

\noindent\textbf{Icentia11k pretraining and downstream splits:}
We use Icentia11k \cite{tan2019icentia11k} for long-context single-lead rhythm evaluation. The dataset is sampled at 250 Hz and contains 11,000 patients. We split patients into 8,800 SSL pretraining patients, 1,100 supervised training patients, 550 validation patients, and 550 test patients. All splits are patient-level.

\noindent\textbf{Icentia preprocessing and labeling:}
ECG records are bandpass filtered between 0.5 and 40 Hz using a zero-phase Butterworth filter and normalized by per-window z-scoring. We use 60-second windows, corresponding to 15,000 samples. Rhythm annotations include normal sinus rhythm (NSR), atrial fibrillation (AFib), and atrial flutter (AFL). Those annotations are converted into non-overlapping intervals. A window is retained only if the annotated rhythm coverage and dominant rhythm occupancy both satisfy a 90\% purity threshold. Labels are binarized as N versus AFib/AFL, with AFib and AFL mapped to the positive class. Beat-synchronous models utilize labeled beat annotations as R-peak locations, keep valid beats (including normal and ectopic beats), and exclude unclassified beats.

\noindent\textbf{Icentia evaluation manifests:}
To ensure fair comparison, validation and test windows are generated once as frozen manifests and reused for all tokenizers. Thus, fixed patching, Tok1, and Tok3 are evaluated on the same patients, records, window start times, and labels. The final validation set contains 1,088 windows, with 83 AFib/AFL positives (7.6\%) and 1,005 N windows (92.4\%). The test set contains 1,082 windows, with 80 AFib/AFL positives (7.4\%) and 1,002 N windows (92.6\%). During supervised training, windows are sampled with a balanced class probability (i.e., $p(\mathrm{AFib/AFL})=0.5$), whereas validation and test sets use the natural class prevalence.

\noindent\textbf{Icentia token counts:}
For fixed patching, $p=160$ gives 93 fixed tokens per 60-second window. For beat-synchronous tokenization, the token count varies with heart rate. Across valid 60-second Icentia windows, the mean number of beats is 68.1, the median is 67, the interquartile range is 59--76, and the 95th percentile is 93. We therefore set the maximum beat sequence length to 93, which covers 95.4\% of windows.

\noindent\textbf{Icentia downstream training and metrics:}
The Icentia downstream task is binary classification of N versus AFib/AFL. Models are trained with cross-entropy loss using AdamW, with learning rate $10^{-4}$ for encoder parameters, $10^{-3}$ for the classifier, and weight decay $10^{-4}$. Each model is trained for 10 epochs with batch size 64, and the best checkpoint is selected by validation AUPRC. We report AUROC and AUPRC on the held-out test set, emphasizing AUPRC because AFib/AFL windows are rare \cite{saito2015precision}.

\noindent\textbf{Computational efficiency evaluation:}
Computational efficiency for fixed patching and Tok1 was measured in FP32 on an NVIDIA GeForce RTX 4090 at batch size 64 using 2,163 usable PTB-XL fold-10 ECGs and 1,623 windows sampled from held-out Icentia11k test patients; timing includes neural tokenization, Transformer encoding, pooling, and classification, but excludes preprocessing and beat extraction.

\noindent\textbf{Data and code availability:}
The data used in this study are publicly available from MIMIC-IV-ECG \cite{PhysioNet-mimic-iv-ecg-1.0}, PTB-XL \cite{PhysioNet-ptb-xl-1.0.3}, and Icentia11k \cite{tan2019icentia11k}. The code and pretrained models are publicly available at \url{https://github.com/muha-0/beat-synchronous-tokenizer}.

\section{Results and Discussion}
\label{sec:results}


\begin{table}[!h]
\caption{PTB-XL five-superclass classification after MIMIC-IV-ECG masked pretraining. Results are mean $\pm$ standard deviation over five runs.\vspace{.5em}}
\label{tab:ptbxl_results}
\centering
\small
\setlength{\tabcolsep}{2.5pt}
\begin{tabular}{@{}lccc@{}}
\toprule
\textbf{Tokenizer} & \textbf{Tokens} & \textbf{Macro AUROC} & \textbf{Macro AUPRC} \\
\midrule
Fixed $p=50$  & 100 & $0.8903 \pm 0.0014$ & $\mathbf{0.7419 \pm 0.0025}$ \\
Fixed $p=100$ & 50  & $0.8858 \pm 0.0007$ & $0.7345 \pm 0.0043$ \\
Fixed $p=250$ & 20  & $0.8717 \pm 0.0008$ & $0.7033 \pm 0.0017$ \\
Fixed $p=500$ & 10  & $0.8479 \pm 0.0019$ & $0.6530 \pm 0.0055$ \\
\midrule
Tok1 & 11.2 avg. & $\mathbf{0.8945 \pm 0.0012}$ & $0.7414 \pm 0.0037$ \\
Tok2 & 11.2 avg. & $0.8276 \pm 0.0034$ & $0.6328 \pm 0.0072$ \\
Tok3 & 11.2 avg. & $0.8928 \pm 0.0015$ & $0.7399 \pm 0.0039$ \\
\bottomrule
\end{tabular}
\end{table}


\noindent\textbf{PTB-XL diagnostic classification:}
Table~\ref{tab:ptbxl_results} shows the main 12-lead diagnostic transfer results on PTB-XL. Among fixed temporal patch tokenizers, the finest patch size, $p=50$, provides the best performance, achieving $0.8903$ macro AUROC and $0.7419$ macro AUPRC. Performance decreases as the fixed patch size increases, indicating that coarse fixed patches lose diagnostically useful waveform detail. In contrast, Tok1 achieves the highest mean macro AUROC ($0.8945$) and nearly matches the best fixed-patch macro AUPRC, with a negligible absolute difference from fixed $p=50$. Tok3 also performs close to fixed $p=50$, reaching $0.8928$ macro AUROC and $0.7399$ macro AUPRC.

\begin{table}[!h]
\caption{Model-side inference efficiency in FP32 on an NVIDIA GeForce RTX 4090 at batch size 64. Tokens are mean real/padded sequence lengths.\vspace{.5em}}
\label{tab:efficiency}
\centering
\small
\setlength{\tabcolsep}{2.5pt}
\begin{tabular}{@{}llccc@{}}
\toprule
\textbf{Dataset} & \textbf{Tokenizer} & \textbf{Tokens} &
\textbf{ms/sample} & \textbf{Memory (MB)} \\
\midrule
PTB-XL & Fixed $p=50$  & 100/100   & 0.0225 & 93.8 \\
       & Tok1          & 11.2/20.1 & 0.0151 & 72.8 \\
\midrule
Icentia11k & Fixed $p=160$ & 93/93     & 0.0298 & 84.7 \\
           & Tok1          & 68.1/92.9 & 0.0308 & 100.9 \\
\bottomrule
\end{tabular}
\end{table}

\noindent\textbf{Token efficiency:}
The PTB-XL results highlight the main efficiency advantage of beat-synchronous tokenization. Fixed $p=50$ uses 100 tokens for each 10-second ECG, whereas Tok1 and Tok3 use 11.2 beat tokens on average. Thus, the beat-synchronous models obtain comparable or slightly better AUROC and nearly identical AUPRC while reducing the average sequence length by almost an order of magnitude. This matters for Transformer-based ECG modeling because self-attention has quadratic complexity in the number of tokens \cite{vaswani2017attention}. Although fixed $p=500$ uses a similar number of tokens to the beat-synchronous models, its performance is much lower, suggesting that token count alone does not explain the result. Table~\ref{tab:efficiency} reports the corresponding model-side inference measurements. On PTB-XL, Tok1 reduces amortized inference time by 33.0\% and peak GPU memory by 22.4\% relative to fixed $p=50$. On Icentia11k, the reduction in real tokens is largely eliminated by dense batch padding, resulting in comparable inference time without a memory advantage. These results highlight that padding strategy is an important factor in whether token-count reductions translate into practical computational savings.

\noindent\textbf{Beat alignment alone is not sufficient:}
Tok2 performs substantially worse than Tok1 and Tok3, despite using the same beat boundaries. This suggests that the benefit of beat-synchronous tokenization depends on how each beat is encoded. Adaptive pooling compresses variable-length beats into a single token after a convolutional front end, but this aggressive compression appears to discard morphology needed for downstream diagnosis. In contrast, Tok1 and Tok3 resample each cardiac cycle to a fixed length before convolutional embedding, preserving more within-beat structure. Therefore, the main conclusion is not simply that ``beats are better,'' but that beat-synchronous tokens must preserve morphology and, when relevant, timing information.


\begin{table}[!h]
\caption{Icentia11k 60-second N vs.\ AFib/AFL classification after patient-level contrastive pretraining. Results are mean $\pm$ standard deviation over five runs.\vspace{.5em}}
\label{tab:icentia_results}
\centering
\small
\setlength{\tabcolsep}{2.5pt}
\begin{tabular}{@{}lccc@{}}
\toprule
\textbf{Tokenizer} & \textbf{Tokens} & \textbf{AUROC} & \textbf{AUPRC} \\
\midrule
Fixed $p=160$ & 93 & $\mathbf{0.9888 \pm 0.0031}$ & $0.8514 \pm 0.0676$ \\
Tok1 & 68.1 avg. & $0.9715 \pm 0.0050$ & $0.8514 \pm 0.0076$ \\
Tok3 & 68.1 avg. & $0.9669 \pm 0.0065$ & $\mathbf{0.8515 \pm 0.0202}$ \\
\bottomrule
\end{tabular}
\end{table}

\noindent\textbf{Icentia11k long-context rhythm classification:}
Table~\ref{tab:icentia_results} reports the 60-second Icentia11k rhythm classification results. Fixed patching obtains the highest AUROC, $0.9888$, indicating stronger overall ranking performance in this single-lead rhythm setting. However, AFib/AFL windows are rare in the held-out test set, comprising only 7.4\% of examples, making AUPRC especially important \cite{saito2015precision}. Under this metric, Tok1 and Tok3 achieve essentially the same mean AUPRC as fixed patching while using fewer tokens on average. Tok1 reaches $0.8514$ AUPRC, and Tok3 obtains the highest mean AUPRC, $0.8515$, although the difference is negligible. These results suggest that beat-synchronous tokenization remains competitive for imbalanced long-context rhythm classification.

\noindent\textbf{Stability under class imbalance:}
The Icentia11k results also show a notable difference in run-to-run variability. Fixed patching has a large AUPRC standard deviation of $0.0676$, while Tok1 and Tok3 have smaller standard deviations of $0.0076$ and $0.0202$, respectively. Since validation and test windows are frozen and identical across tokenizers, this difference is not due to changing evaluation examples. A plausible explanation is that beat-synchronous tokenization imposes a stronger rhythm-level inductive bias by presenting the model with cardiac-cycle units rather than arbitrary fixed patches. This may be helpful when the positive class is rare and precision-recall performance is sensitive to a small number of difficult examples.

\noindent\textbf{Overall interpretation:}
Across both settings, beat-\hspace{0pt}synchronous tokenization provides a favorable performance--efficiency tradeoff, but it is not uniformly superior on every metric. On PTB-XL, Tok1 and Tok3 match the strongest fixed-patch baseline with far fewer tokens. On Icentia11k, beat-synchronous tokenizers match fixed-patch AUPRC and are more stable across runs, but fixed patching achieves higher AUROC. These findings support a balanced conclusion: cardiac-cycle alignment is a useful inductive bias for ECG Transformers, especially for compact sequence modeling, but the design of the beat encoder is critical. Morphology-preserving beat representations work well, whereas overly compressed beat tokens lose important diagnostic details.

\section{Conclusion}
\label{sec:conclusion}

We studied two tokenization strategies for Transformer-based ECG representation learning, comparing fixed temporal patches with beat-synchronous tokenizers under matched pretraining and downstream evaluation protocols. Across 12-lead PTB-XL diagnostic classification and 60-second Icentia11k rhythm classification, beat-synchronous tokenization achieved competitive performance with shorter, physiologically meaningful token sequences. The strongest beat-based variants, Tok1 and Tok3, matched fine fixed-patch performance on PTB-XL while using substantially fewer tokens, and achieved comparable AUPRC with lower run-to-run variability on imbalanced Icentia11k rhythm classification. However, the poor performance of the adaptive-pooling variant of beat-synchronous tokenization shows that beat alignment alone is not sufficient: beat-token design must preserve morphology and, when relevant, timing information. A limitation is the dependence on reliable R-peak localization; we did not evaluate sensitivity to missed or shifted R-peaks, and noisy recordings may therefore require robust beat detection or fallback fixed-patch tokenization. These results suggest that beat-synchronous tokenization is a promising token-efficient alternative to fixed temporal patching. Future work should explore alternative beat extraction strategies, broader use of R--R timing information, stronger beat encoders, and evaluation across additional ECG tasks.

\balance
\small
\bibliographystyle{IEEEbib}
\bibliography{refs}

\end{document}